\pdfoutput=1

\documentclass[11pt]{article}

\usepackage[]{acl}

\usepackage{times}
\usepackage{latexsym}
\usepackage{ulem}
\useunder{\uline}{\ul}{}
\usepackage[T1]{fontenc}

\usepackage[utf8]{inputenc}

\usepackage{microtype}

\usepackage{inconsolata}
\usepackage{amsmath}
\usepackage{mathrsfs}
\usepackage{algorithm}
\usepackage{tcolorbox}
\usepackage{booktabs}
\usepackage{multirow}
\usepackage{graphicx}
\usepackage{svg}
\usepackage{subcaption} 

\title{MoL-RL: Distilling Multi-Step Environmental Feedback into LLMs for Feedback-Independent Reasoning}

\author{
    \textbf{Kang Yang$^{1}$, Jingxue Chen$^{2}$, Qingkun Tang$^2$\thanks{\,Qingkun Tang is the corresponding author.}, Tianxiang Zhang$^2$, Qianchun Lu$^2$}\\\\
    $^1$Dalian University of Technology, Dalian, China\\
    $^2$Wired Product Operation Division, ZTE Corporation, Nanjing, China\\
    \small {\textbf{E-mail:} chen.jingxue@zte.com.cn, tang.qingkun@zte.com.cn}
}

\begin{document}
\maketitle
\begin{abstract}
Large language models (LLMs) face significant challenges in effectively leveraging sequential environmental feedback (EF) signals, such as natural language evaluations, for feedback-independent chain-of-thought (CoT) reasoning. Existing approaches either convert EF into scalar rewards, losing rich contextual information, or employ refinement datasets, failing to exploit the multi-step and discrete nature of EF interactions. To address these limitations, we propose MoL-RL, a novel training paradigm that integrates multi-step EF signals into LLMs through a dual-objective optimization framework. Our method combines MoL (Mixture-of-Losses) continual training, which decouples domain-specific EF signals (optimized via cross-entropy loss) and general language capabilities (preserved via Kullback-Leibler divergence), with GRPO-based post-training to distill sequential EF interactions into single-step inferences. This synergy enables robust feedback-independent reasoning without relying on external feedback loops. Experimental results on mathematical reasoning (MATH-500, AIME24/AIME25) and code generation (CodeAgent-Test) benchmarks demonstrate that MoL-RL achieves state-of-the-art performance with the Qwen3-8B model, while maintaining strong generalization across model scales (Qwen3-4B). This work provides a promising approach for leveraging multi-step textual feedback to enhance LLMs' reasoning capabilities in diverse domains.
\end{abstract}

\begin{figure*}[h]
\centering
\includegraphics[width=1.0\textwidth, trim=0pt 0pt 0pt 5pt, clip]{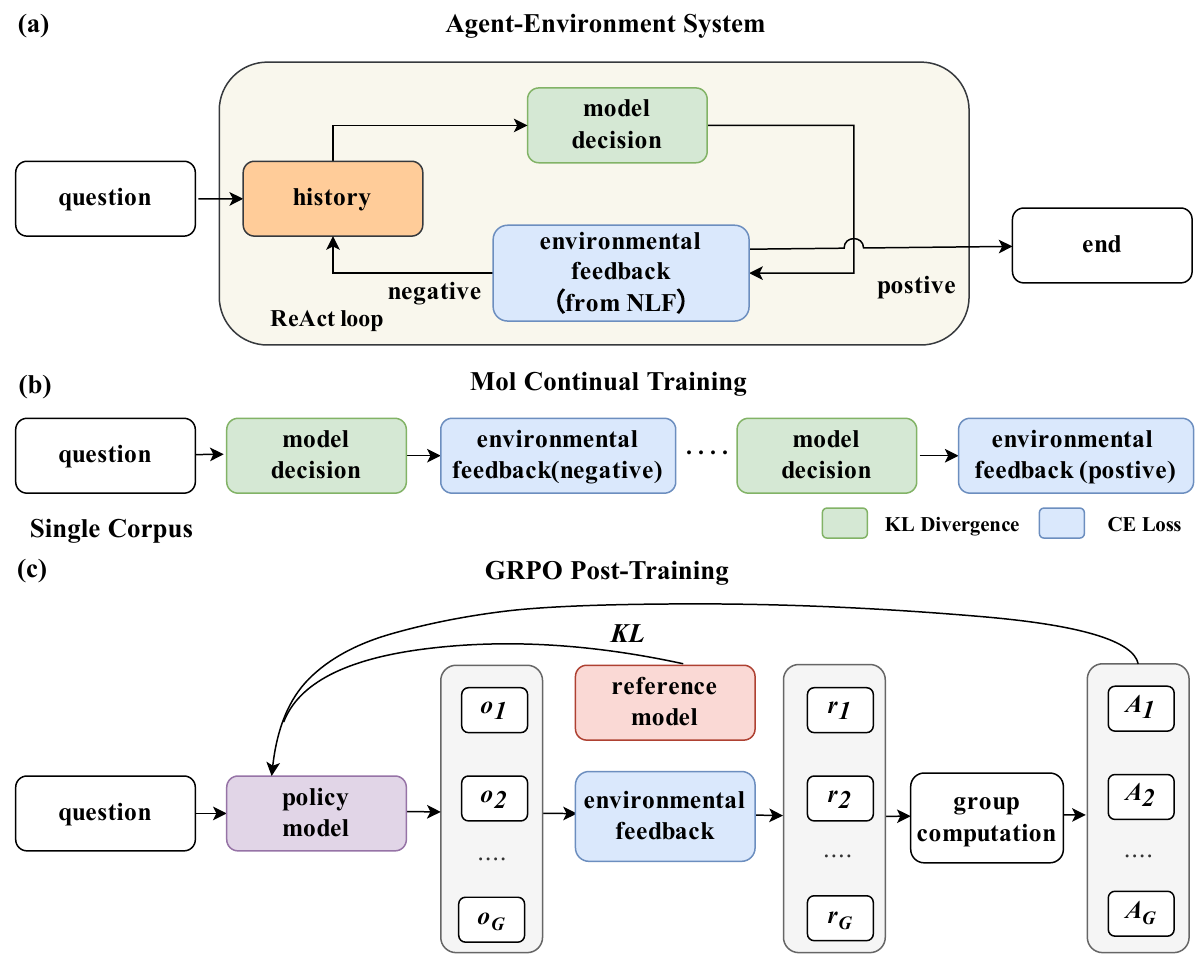} 
\caption{Overview of the MoL-RL training process. (a) Illustration of the agent-environment interaction system constructed using NLF. (b) Depicts the multi-stage training process under the MoL paradigm. The model's self-generated reasoning trajectories are optimized through KL divergence, while EF signals are processed via CE loss. (c) Schematic of the GRPO-based RL implementation. The final model output is supervised by a reward function derived directly from EF.}
\label{figure0}
\end{figure*}

\section{Introduction}
Reinforcement learning (RL) algorithms that use numerical or deterministic feedback signals have demonstrated effectiveness in improving the chain-of-thought (CoT) reasoning capabilities of large language models (LLMs) \cite{guo2025deepseek}. Currently, natural language feedback (NLF) mechanisms have been empirically validated to provide superior auxiliary benefits during model inference \cite{liu2025inference}, which can be conceptualized as environmental feedback (EF) signals generated during agent-environment interactions \cite{tang2024worldcoder}. However, the effective integration of general textual EF into model training remains an unresolved challenge in the field. Most existing approaches either convert such feedback into numerical representations \cite{ lightman2023let,whitehouse2025j1}, thereby losing the rich informational content inherent in NLF, or construct refinement datasets for fine-tuning model capabilities \cite{chen2024learning, wang2025critique}. While these methods fundamentally overlook the discrete and multi-step nature of EF signals \cite{madaan2023self}, where single-step numerical conversions or isolated refinement processes are inherently insufficient to fully exploit their sequential information and compound effects across interaction steps. 

To address the limitations of traditional approaches and effectively leverage multi-step EF signals to enhance feedback-independent CoT reasoning capabilities, we propose MoL-RL, a novel training methodology. Here, the acronym MoL (Mixture-of-Losses) denotes a continual pre-training approach that decouples the loss functions for domain-specific and general corpora \cite{chen2025mol}. Specifically, domain-specific corpora are optimized using cross-entropy (CE) loss, while general corpora employ Kullback-Leibler (KL) divergence. This architecture prevents catastrophic forgetting of general capabilities while enabling knowledge absorption from domain-specific contexts. In our method, EF is viewed as pending-absorption domain knowledge, while model decisions constitute general corpora. This dual-objective optimization ensures continuous compression of critical EF information while maintaining general language proficiency. After continual pre-training, we further implement Group Relative Policy Optimization (GRPO) \cite{shao2024deepseekmath} based RL algorithm to synergistically activate both the model's inherent general capabilities and the compressed EF knowledge, thereby distilling sequential interactions into a single-step direct output and inducing emergent feedback-independent reasoning capabilities.

Our contributions include three key innovations:\\
(1) We demonstrate the effectiveness of MoL in agent-environment interaction systems for LLM continual learning. This approach prevents catastrophic forgetting while continuously compressing essential EF information into the model, establishing a solid foundation for subsequent RL-driven feedback-independent reasoning.\\
(2) Our experiments demonstrate that the GRPO-based RL algorithm effectively harnesses multi-step EF signals to activate emergent CoT reasoning capabilities. Ablation studies reveal that the sequential integration of EF through MoL continual training significantly enhances the model's responsiveness to RL, as evidenced by the positive correlation between MoL training steps and RL phase reward improvement potential.\\ 
(3) Our MoL-RL method establishes a generalizable paradigm for continual learning in agent-environment systems, demonstrating broad applicability to enhancement of reasoning capabilities and optimization of code generation optimization.

\section{Related Work}
\paragraph{Learning from NLF} NLF has emerged as a powerful mechanism to enhance reasoning capabilities in LLMs \cite{liu2025inference}. By providing detailed and targeted insights into model output defects through textual evaluations, NLF enables precise assessment and optimization of response quality. Two primary paradigms have been explored in the existing literature: (1) numerical reward conversion approaches that transform NLF into scalar signals for RL applications \cite{lightman2023let,rafailov2024scaling,whitehouse2025j1}, and (2) direct learning frameworks that leverage NLF through fine-tuning processes to improve generation quality \cite{chen2024learning,wang2025critique}. Recent advances have introduced hybrid RL approaches that synergistically utilize both raw NLF and converted numerical rewards \cite{zhang2025critique}. However, our MoL-RL methodology innovates by establishing an end-to-end training paradigm that directly incorporates multi-step NLF. This approach fundamentally differs from previous work by treating NLF as a continual process of knowledge absorption, rather than a transient optimization signal, thereby enabling more sustainable and generalizable reasoning capability development in LLMs.


\begin{figure}
\centering
\includegraphics[width=1.0\columnwidth, trim=5pt 5pt 0pt 5pt, clip]{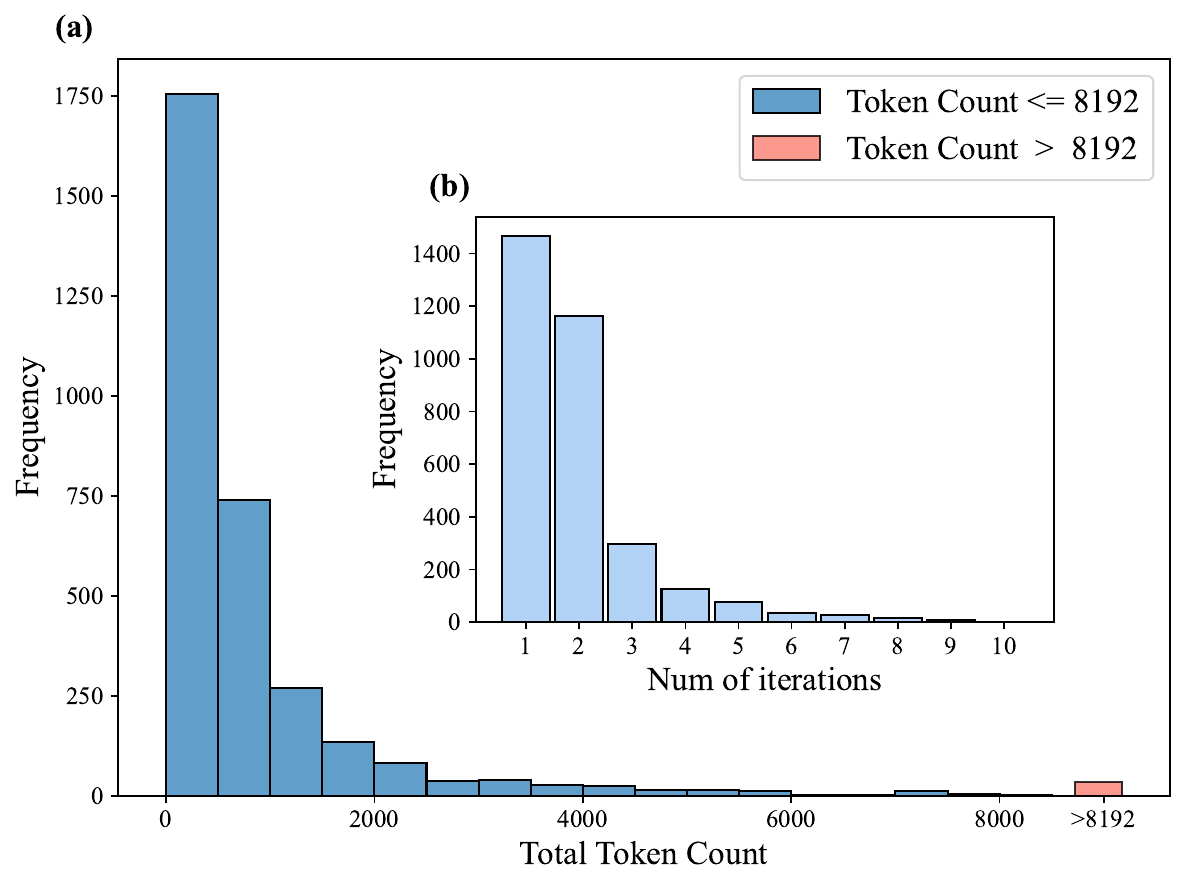}
\caption{(a) Token count distribution across multi-turn dialogues in the training data. (b) Statistical distribution of dialogue turn counts aligned with EF interaction frequencies.}
\label{figure1}
\end{figure}

\paragraph{MoL Training Approach} The MoL training approach employs a dual-loss optimization for LLMs continual pre-training, applying CE loss to domain-specific corpora while utilizing KL divergence relative to a base model for general corpora \cite{chen2025mol}. In the context of multi-step agent-environment interactions, we identify a dual relationship between observational data (EF) and cognitive processes (model decision) \cite{leung2025llm}, which directly parallels the domain-general dichotomy in traditional corpus training. This conceptual alignment motivates our novel application of MoL to sequential interactive learning, where environmental observations are treated as domain-specific knowledge sources, while model-generated thoughts constitute the general knowledge repository. This architectural extension not only preserves the advantages of original MoL methodology, but also establishes a fundamental mechanism for sequential knowledge absorption in agent-environment interaction systems.

\section{Methods}
\subsection{Problem Modeling}
\label{subsec:problem_modeling}
\paragraph{From NLF to multi-step EF} NLF can be formalized as a sequential interaction pattern, 
\begin{align}
    Q{\rightarrow}A{\rightarrow}C,
\end{align}
where $Q$ denotes a system prompt together with a query, $A$ represents the model's response, and $C$ explicitly corresponds to the NLF that evaluates the quality of $A$. In the context of LLMs, this single-step evaluation mechanism naturally extends to a multi-step process until an optimized answer is achieved (see Figure \ref{figure0} (a) and (b) for details), which forms the sequence, 
\begin{align}
Q{\rightarrow}A_1{\rightarrow}C_1{\rightarrow}...{\rightarrow}A_n{\rightarrow}C_n{\rightarrow}A_{n+1}{\rightarrow}C_{n+1}.
\label{sequence}
\end{align} 
This recursive structure parallels the agent-environment interaction paradigm in RL, where $A$ corresponds to the model's reasoning decisions and $C$ represents EF. This conceptual alignment enables a unified approach for modeling sequential interactions in LLM training. For generalizability, all subsequent discussions will adopt the agent-environment interaction framework as the default paradigm for modeling sequential interactions in LLM training.

\begin{table*}[!htbp]

\centering 
\begin{tabular}{cccccc}
\hline
\textbf{Model} & \textbf{MATH-500} & \textbf{AIME24} & \textbf{AIME25} & \begin{tabular}[c]{@{}c@{}}\textbf{CodeAgent-Test}\end{tabular} &\begin{tabular}[c]{@{}c@{}}\textbf{CodeAgent-Test}\\ (non-thinking)\end{tabular}\\ \hline
Qwen3-8B  & {\ul 96.60}    & {\ul 76.67}          & {\ul 66.67}          & \textbf{83.00} &  \textbf{37.50} \\
+MoL      & \textbf{97.40} & {\ul 76.67}          & \textbf{68.89} & {\ul 76.00}  &{\ul 37.00} \\
+CE       & 95.80          & 72.22          & 65.56          & 32.00 & 27.50 \\
+CE\_NoKL & \textbf{97.40} & \textbf{80.00} & 57.78          & 4.50 & 0 \\
\begin{tabular}[c]{@{}c@{}}+CE (EETB)\end{tabular}   & {\ul96.60} &  75.56  &   65.56 & 73.00 & 30.50 \\ 
\hline
\end{tabular}

\caption{Comparison of model performance across different training methods on four benchmark datasets (MATH-500, AIME24, AIME25, CodeAgent-Test). Values represent accuracy/success rates. "+CE" replaces KL divergence in MoL with CE; "+CE\_NoKL" disables KL divergence for model decisions; "+CE (EETB)" excludes empty "think" blocks in the pure CE loss. \textbf{Bold values} indicate the best performance, while \underline{underlined values} indicate the second-best performance on each dataset.}
\label{Mol}
\end{table*}

\paragraph{From multi-step EF to CoT}
In the multi-step feedback framework, we propose distilling the iterative sequence, 
\begin{align}
A_1{\rightarrow}C_1{\rightarrow}...{\rightarrow}A_n{\rightarrow}C_n
\end{align}
in Equation \ref{sequence} into a compressed reasoning chain $R$. Assuming that this distillation process successfully captures the causal relationships between EF and optimal decisions, we can realize feedback-independent reasoning through the transformation $Q{\rightarrow}R{\rightarrow}A$. This would enable the model to autonomously synthesize critical EF interactions into a single-step inference without requiring external feedback loops. However, existing methods rely on prompt engineering to elicit post-hoc reasoning traces from LLMs \cite{kang2025c3ot}, such approaches face two critical limitations: (1) the quality of transformed reasoning chains is highly sensitive to prompt design, and (2) there exists no direct mechanism to evaluate the fidelity of the compressed reasoning. To address these challenges, we propose the MoL-RL methodology, which synergistically combines MoL continual training and GRPO-based post-training to realize end-to-end absorption of multi-step EF, thereby enabling feedback-independent reasoning capabilities through sequential knowledge compression.

\subsection{Training Pipeline}
\label{subsec:framework}
The MoL-RL methodology establishes a two-phase training pipeline that synergistically integrates sequential knowledge compression and policy refinement (see Figure \ref{figure0} (b) and (c) for architecture overview). 

The first phase employs MoL continual training to compress multi-step EF signals into the model's knowledge base through dual-objective optimization: (1) CE loss absorbs domain-specific EF signals from iterative interactions, and (2) KL divergence preserves general language capabilities by treating model-generated reasoning as general knowledge.

The second phase applies GRPO-based post-training to distill the compressed knowledge into feedback-independent reasoning capabilities. The combined MoL-RL pipeline enables the model to autonomously synthesize multi-step EF interactions into single-step inferences, achieving robust generalization across diverse problem domains without external feedback loops.

\begin{table*}[!htbp]
\centering 
\begin{tabular}{ccccc}
\hline
\textbf{Model}       & \textbf{MATH-500} & \textbf{AIME24} & \textbf{AIME25} & \textbf{CodeAgent-Test} \\ \hline
Qwen3-4B &
  \begin{tabular}[c]{@{}c@{}}97.00\\ (official)\end{tabular} &
  \begin{tabular}[c]{@{}c@{}}{\ul 73.80}\\ (official)\end{tabular} &
  \begin{tabular}[c]{@{}c@{}}65.60\\ (official)\end{tabular} &
  \begin{tabular}[c]{@{}c@{}}79.00\end{tabular} \\
+MoL         & \textbf{97.40}              & \textbf{76.67}  & 66.67           & 76.50                    \\
+MoL+GRPO    & {\ul 97.20}           & \textbf{76.67}  & {\ul 66.67}     & \textbf{84.00}           \\
+MoL+Dr.GRPO &    \textbf{97.40}             & 73.33           & \textbf{74.45}  & {\ul 82.00}              \\ \hline
Qwen3-8B             & 96.60             & 76.67           & {\ul 66.67}           & 83.00                    \\
+Mol         & {\ul 97.40}       & 76.67           & \textbf{68.89}  & 76.00                    \\
+MoL+GRPO    & \textbf{98.00}    & {\ul 77.78}     & {\ul 66.67}           & {\ul 85.50}              \\
+MoL+Dr.GRPO & 97.00             & \textbf{80.00}  & \textbf{68.89}  & \textbf{87.00}           \\ \hline
\end{tabular}
\caption{Performance comparison of different model configurations across four benchmark datasets (MATH-500, AIME24, AIME25, CodeAgent-Test). Results demonstrate the effectiveness of MoL-RL's two-phase training pipeline: (1) "+MoL" represents continual pre-training with domain-specific KL/CE loss; (2) "+GRPO"/"+Dr.GRPO" indicates post-training with standard/grouped relative policy optimization. Values in parentheses for Qwen3-4B denote official released metrics. \textbf{Bold} and \underline{underlined} values highlight top-2 performances per dataset. The Qwen3-8B+MoL+Dr.GRPO variant achieves the highest accuracy on CodeAgent-Test (87.00\%) and maintains strong performance across almost tasks.}
\label{Mol+GRPO}
\end{table*}


\subsection{MoL continual training}
\label{subsec:MoL}
Our implementation of the MoL approach differs from the conventional methodology described in prior work \cite{chen2025mol}. We propose a relaxed formulation that accommodates proportional mixing of domain-specific and general tokens within a single training batch. This flexibility ensures the reuse of established MoL optimization principles while enabling a more efficient utilization of heterogeneous data streams in agent-environment interaction systems.

In the context of sequential decision making, our MoL formulation addresses two critical challenges inherent in EF learning. First, the model must acquire a comprehensive understanding of EF signals in various decision strategies to optimize policy selection. This can be conceptualized as an extreme scenario: if the model pre-knows all potential EF outcomes for every possible action, it may theoretically achieve optimal decisions without external feedback, effectively transforming sequential interactions into single-step inferences. Second, the process of EF knowledge absorption inevitably incurs generalization risks. Analogously to incremental knowledge infusion in continual learning, naively integrating EF signals without regularization would lead to catastrophic forgetting of foundational language capabilities.

Thus, an MoL algorithm tailored for agent-environment interaction systems is feasible and necessary. By treating $A$ as general knowledge optimized via KL divergence and $C$ as domain-specific knowledge optimized via CE loss (Equation \ref{sequence}), the framework achieves effective sequential knowledge integration. For instance, in a code generation task where the model iteratively refines a solution based on execution feedback, intermediate reasoning steps $(A_1, A_2,...,A_{n+1})$ are preserved as general knowledge via KL divergence to maintain foundational coding skills, while execution messages $(C_1, C_2,...,C_{n+1})$ are treated as domain-specific signals to refine task-specific accuracy (see Example \ref{MoL_example} for a concrete illustration). The loss function $\mathcal{L}$ for each sequence is formulated as follows:
\begin{align}
\mathcal{L} &= \sum_{i=1}^{n+1}\sum_j^{|A_i|}\mathcal{L}_{\text{KL}}(A_{i,j}) \notag \\&+ \sum_{i=1}^{n+1}\sum_j^{|C_i|}\mathcal{L}_\text{CE}(C_{i,j})
\label{loss_fun}
\end{align}
where the summation over $i$ spans the interaction steps, $j$ spans each token, and other parameters align with Equation \ref{sequence}. 

Two alternative approaches warrant consideration: (1) A naive application of CE loss to $A$ risks propagating suboptimal or erroneous decisions through the training process, as intermediate reasoning steps may contain inherent flaws. (2) In contrast, complete disregard of $A$ during training would compromise foundational language capabilities through catastrophic forgetting, as previously discussed. Through systematic empirical validation, our work demonstrates that the proposed MoL-based implementation achieves optimal trade-offs between knowledge absorption and capability preservation. This approach not only mitigates the limitations of competing strategies, but also establishes a robust method for sequential knowledge compression in interactive learning systems.

\begin{table*}[!htbp]
\centering 
\begin{tabular}{cccccc}
\hline
\textbf{Model}                 & \textbf{MATH-500} & \textbf{AIME24} & \textbf{AIME25} & \textbf{CodeAgent-Test} \\ \hline
Qwen3-8B                 & 96.60 & 76.67 & 66.67 & 83.00 \\
+MoL+Dr.GRPO (multi-value reward)          & {\ul 97.00}       & \textbf{80.00}  & {\ul 68.89}     & \textbf{87.00} \\
+MoL+Dr.GRPO (binary reward) & \textbf{97.20}    & {\ul 78.89}     & \textbf{70.00}  & {\ul 86.50} \\ \hline
\end{tabular}

\caption{Comparison of model performance under different reward formulations (multi-value vs. binary reward) in the MoL-RL approach. The Qwen3-8B+MoL+Dr.GRPO variant demonstrates consistent performance across both reward types on four benchmark datasets (MATH-500, AIME24, AIME25, CodeAgent-Test).}
\label{reward}
\end{table*}

\subsection{GRPO-based post-training}
\label{subsec:grpo}
GRPO is a representative RL algorithm that has demonstrated efficacy in activating CoT reasoning capabilities. And Dr.GRPO, a typical variant of GRPO is proposed to eliminate redundant output while preserving reasoning performance\cite{liu2025understanding}, aligning with our objective of distilling sequential interactions into compact, feedback-independent reasoning chains. In this work, we systematically compare the effectiveness of both GRPO and Dr.GRPO in achieving optimal trade-offs between reasoning fidelity and output efficiency.

\section{Experiments}
\label{experiments}
Our objective is to validate the effectiveness of the MoL-RL methodology through empirical evaluation on agent-environment interaction datasets. Specifically, we first perform continual training of MoL on the interaction datasets, leveraging KL divergence and  CE loss to learn the model's reasoning capabilities and EF signals, respectively. Subsequently, we apply GRPO to perform RL on the final EF, further refining the knowledge acquired during the MoL training phase. Finally, we evaluate the trained model's performance in mathematical reasoning and environment-agnostic inference to comprehensively validate the effectiveness of the proposed methodology.

\paragraph{Datasets} Our training data is derived from the high-quality CodeAgent-Traces dataset \cite{smolagents_githubs}, which captures sequential agent-environment interactions where CodeAgents solve complex tasks requiring program synthesis, including mathematical problem-solving and code generation. Each sample consists of iterative reasoning trajectories ($A_1$,$C_2$,...,$A_n$,$C_n$) where the model generates code-based solutions ($A$) and receives EF ($C$) through tool execution. We carefully filter out samples requiring external API calls, retaining only those relying on local computational environments. To ensure the length of multi-turn dialogues remains within a learnable range, we further process the dataset by clearing the "think" blocks in the decision portion of the reasoning chains for CoT models, while non-CoT models are augmented with empty "think" blocks ("<think>\textbackslash n\textbackslash n</think>") to maintain structural consistency. This preprocessing step effectively reduces redundancy while preserving critical interaction signals. The resulting token count distribution and dialogue turn statistics after this operation are presented in Figure \ref{figure1}. We randomly selected 100 samples as the validation set, an additional 200 samples as the test set, and a final training set of approximately 3,000 curated interaction sequences.

\paragraph{Benchmarks} To evaluate the mathematical reasoning capabilities of the model trained using the MoL-RL approach, we utilized four benchmarks: AIME24, AIME25 \cite{aime}, MATH-500 \cite{math500}, and a custom test set named CodeAgent-Test, which consists of the 200 test samples retained from the CodeAgent-Traces dataset. For all benchmarks, we report the pass@1 results. To mitigate the variability of pass@1 results on the smaller datasets, AIME24 and AIME25, we report the average results over three evaluation runs. For the other two benchmarks, we report the results of a single evaluation.

\paragraph{MoL continual training} We select the open-source Qwen3-8B and Qwen3-4B model \cite{yang2025qwen3}, which are specifically designed with a CoT reasoning architecture to support multi-step logical inference, as the base model and implement MoL continual training for agent-environment interactions based on the PEFT framework \cite{peft}. During this process, we employ the Low-rank Adaptation (LoRA) training method\cite{hu2022lora}, setting the rank to 64. Furthermore, throughout the MoL continual training, the model's context length is fixed at 8192 tokens to ensure the effective learning of long input sequences in agent-environment interactions. For completeness, other hyperparameters are detailed in Appendix \ref{MoL_Hyperparameters}.

\begin{figure*}[h]
\centering
\includegraphics[width=1.0\textwidth, trim=10pt 10pt 30pt 20pt, clip]{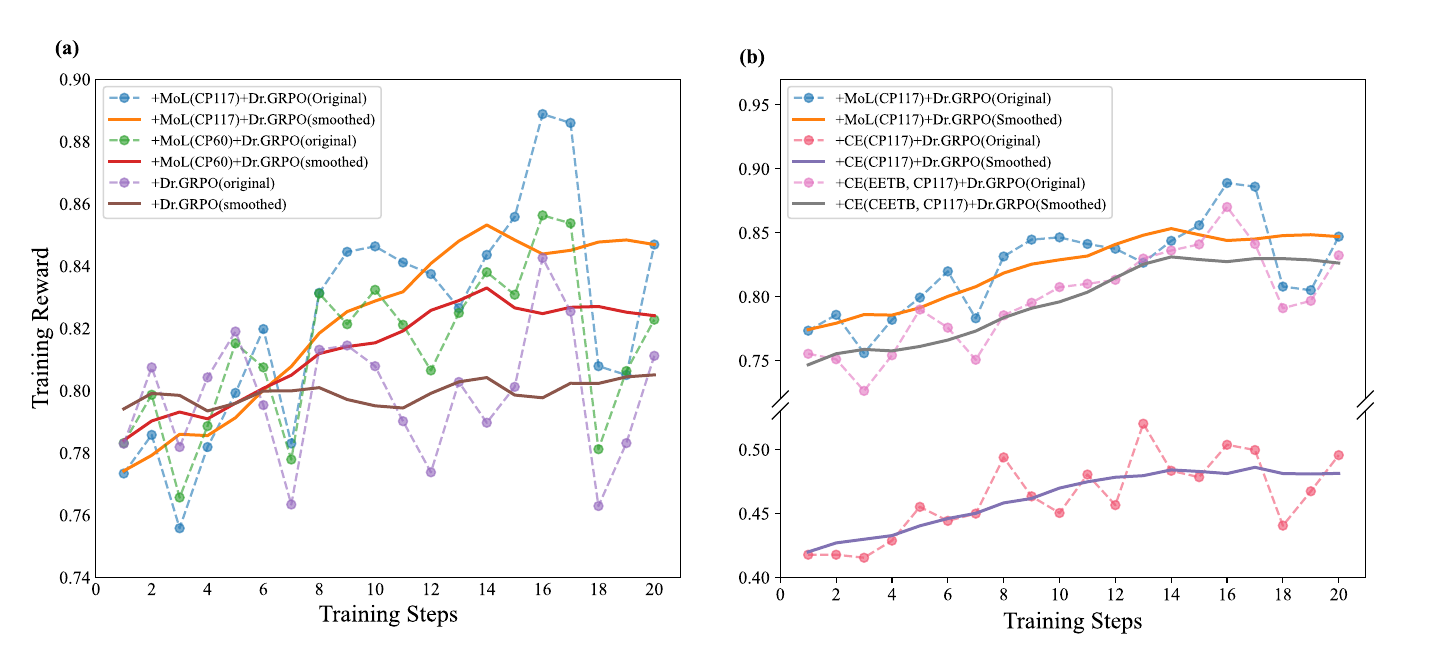} 
\caption{(a) Comparison of reward evolution during Dr.GRPO training on the training set for three representative MoL checkpoints: the base model (0 training steps), checkpoint60 (60 training steps), and checkpoint117 (3 full training epochs). (b) Comparison of reward trends on the training set during Dr.GRPO training for three configurations: (1) MoL training for 3 epochs, (2) pure CE training for 3 epochs, and (3) pure CE training but EETB for 3 epochs. The reward evolution patterns that observed in these configurations are nearly identical.}
\label{figure2}
\end{figure*}

\paragraph{GRPO-based post-training} After MoL continual training, we design a specialized EF signals to conduct GRPO post-training, aiming to enhance the model's environment-agnostic reasoning capabilities. This EF signal is defined as follows. The model is guided through step-by-step reasoning in the system prompt and generates code as the final output. The generated code is executed in a sandbox environment to obtain the execution results, which are then compared against the ground truth. For this type of EF signal, we designed a multi-stage reward function \cite{wu2025recode,pennino2025reasoning} to serve as the GRPO reward. Our reward mechanism evaluates the quality of generated code through three hierarchical criteria: (1) Code Parsing Validity (0.1 points) - awarding partial credit if the model's output can be syntactically parsed without errors; (2) Execution Feasibility (0.4 points) - granting intermediate credit when the code successfully compiles or interprets in the sandbox environment; and (3) Functional Correctness (1.0 point) - full reward is allocated only when the executed code produces results matching the ground truth specification. Additional hyperparameters for the GRPO framework are detailed in Appendix \ref{GRPO_Hyperparameters}.

\paragraph{Evaluation} For the evaluation of all benchmarks, we conduct experiments in thinking mode unless specified. For all models, we adopt the sampling hyperparameters officially recommended for Qwen3 \cite{yang2025qwen3}: a temperature of 0.6, a top-p value of 0.95, a top-k value of 20, and a presence penalty of 1.0. The maximum output length is limited to 30,720 tokens.

\section{Results}
\subsection{Main Results}
We conducted MoL training on the Qwen3-8B model for three epochs and compared its performance against two representative alternative approaches: (1) a variant where all KL divergence terms were replaced with CE loss; (2) a configuration where KL divergence components were entirely disabled (i.e., no KL divergence applied to model-generated reasoning). The experimental results are shown in Table \ref{Mol}. Specifically, the MoL-enhanced Qwen3-8B model achieves the highest accuracy on AIME25 (68.89\%), while maintaining competitive performance on the other benchmarks. This comparative analysis validates the effectiveness of the dual-objective optimization strategy in preserving general language capabilities while effectively absorbing domain-specific EF signals.

\begin{table*}[!htbp]
\centering 
\begin{tabular}{ccccc}
\hline
\textbf{Model} & \textbf{MATH-500} & \textbf{AIME24} & \textbf{AIME25} & \textbf{CodeAgent-Test} \\ \hline
Qwen3-8B               & 96.60          & 76.67          & 66.67          & 83.00          \\
+Dr.GRPO  & {\ul 97.20}    & {\ul 77.78}    & 65.56          & 83.50  \\
+MoL(CP60)+Dr.GRPO  & \textbf{97.40} & {\ul 77.78}    & {\ul 67.78}    & 85.50    \\
+MoL(CP117)+Dr.GRPO & 97.00          & \textbf{80.00} & \textbf{68.89} & \textbf{87.00} \\ 
+CE(CP117)+Dr.GRPO &   96.80      & {\ul 77.78} & 66.89 & 41.00 \\ 
+CE(EETB, CP117)+Dr.GRPO &   96.80     & 76.67 &64.44 & {\ul 86.00} \\
\hline
\end{tabular}

\caption{Ablation study results comparing different model configurations on four benchmark datasets (MATH-500, AIME24, AIME25, CodeAgent-Test). The table highlights the effectiveness of MoL-RL components: (1) "+Dr.GRPO" demonstrates post-training refinement benefits; (2) "+MoL(CP60/CP117)+Dr.GRPO" variants show performance improvements with increasing MoL training steps; (3) "+CE(CP117)+Dr.GRPO" reveals performance degradation due to overfitting. (4) "+CE(EETB, CP117)+Dr.GRPO" " demonstrates the limitations of CE-based optimization in maintaining overall reasoning capabilities, particularly showing a significant performance drop on the complex AIME25 benchmark.  }
\label{ablation}
\end{table*}

We observed a notable performance degradation in the CodeAgent-Test under the "+CE" configuration (Table \ref{Mol}). This discrepancy arises from the CE loss overfitting to empty "think" blocks, thereby inducing a non-thinking response pattern that deviates from the model's native reasoning framework. In contrast, our MoL methodology remains unaffected by this limitation due to the KL divergence component, which inherently avoids pattern-specific learning. However, to rigorously validate the superiority of MoL, we conducted two complementary experiments (as shown in Table \ref{Mol}): (1) a systematic comparison under non-thinking mode, where MoL consistently outperformed CE-based approaches, and (2) a specialized study excluding empty "think" blocks (EETB) from the loss calculation, which further confirmed MoL's robustness across both thinking and non-think paradigms. These empirical validations solidify the methodological advantage of MoL in preserving reasoning fidelity while mitigating pattern-specific overfitting.

We further applied GRPO and Dr.GRPO post-training to the MoL-pretrained Qwen3-8B model, achieving significant performance improvements. As shown in Table \ref{Mol+GRPO}, both GRPO and Dr.GRPO variants demonstrated notable gains in reasoning accuracy compared to the base model. Notably, the Dr.GRPO variant achieved the highest accuracy on CodeAgent-Test (87.00\%) with the Qwen3-8B + MoL + Dr.GRPO configuration, while maintaining strong performance across all benchmark tasks. To validate the generalizability of our training pipeline, we replicated the MoL-RL workflow on the Qwen3-4B model architecture. The results demonstrated consistent improvements in mathematical reasoning and code generation capabilities, with the Qwen3-4B + MoL + Dr.GRPO variant achieving 82.00\% accuracy on CodeAgent-Test (Table \ref{Mol+GRPO}). These findings confirm the robustness of our proposed methodology across different model scales.

The reward function for code execution feedback incorporates configurable parameters, such as the multi-value reward scheme described in Section \ref{experiments}. To evaluate the robustness of our training methodology to reward formulation, we conducted comparative experiments between the baseline multi-value reward (0.1 for partial credit, 0.4 for execution feasibility, 1.0 for functional correctness) and a simplified binary reward scheme (0 for incorrect outcomes, 1 for correct outcomes). As demonstrated in Table \ref{reward}, the GRPO and Dr.GRPO variants exhibited consistent performance across both reward formulations. This insensitivity to reward formulation further validates the generalizability of the MoL-RL approach in sequential decision-making contexts.

\subsection{Ablation Studies and Critical Analysis}
To further investigate the role of MoL in our methodology, we conduct ablation studies using representative intermediate checkpoints: the base model, checkpoint60 (60 training steps) and checkpoint117 (3 full training epochs). Our analysis reveals that as the number of MoL training steps increases, the average reward on the training set exhibits a marginal decline. However, the potential for reward improvement during the RL phase demonstrates a positive correlation with the number of MoL training steps. Specifically, with a limited amount of RL data, the reward exceeds the baseline, as illustrated in Figure \ref{figure2}(a). 

We also compare the performance of the pure CE training approach (3 epochs) followed by RL with that of MoL-based training. As shown in Figure \ref{figure2}(b), the reward trends during RL are nearly identical between these configurations. 

The observed reward discrepancies in the training set are consistently reflected in the benchmark datasets, as evidenced by the Dr.GRPO results with 20 training steps presented in Table \ref{ablation}. This correlation between training dynamics and benchmark performance underscores the importance of reward signal calibration in sequential learning paradigms.

The empirical observations, where the initial performance baseline (determined by the model's foundational capabilities) corresponds to the starting point of the scaling curve and the absorption efficiency of EF governs the ascending trend of the trajectory during RL refinement, align with the scaling behavior observed in post-training. Our findings demonstrate that the MoL approach synergistically optimizes these two critical dimensions:\\
(1) KL divergence preserves the model's general language proficiency, establishing a solid foundation for reasoning performance;\\
(2) CE loss enables sequential EF absorption, directly influencing the steepness of the scaling curve by deepening the model's understanding of the environment, which in turn enhances its ability to make optimal decisions.

\section{Conclusion}
This work introduces MoL-RL, a novel training paradigm that effectively integrates multi-step EF signals into LLMs through a dual-objective optimization approach. By decoupling domain-specific EF absorption (via EF loss) and general language capability preservation (via KL divergence) in MoL continual training, and synergistically distilling sequential interactions into single-step inferences via GRPO-based post-training, MoL-RL achieves state-of-the-art performance on mathematical reasoning and code generation benchmarks. Cross-model evaluations and ablation studies validate its robustness and generalizability, while reward formulation experiments demonstrate its flexibility in decision-making. MoL-RL establishes a promising framework for leveraging textual EF to enhance feedback-independent reasoning, offering a scalable solution for LLMs in complex, interactive domains.

\section*{Limitations}
Although MoL-RL effectively compresses multi-step EF into generalizable reasoning capabilities, its implicit treatment of environment dynamics may limit performance in complex, physically grounded tasks. Unlike world models that explicitly simulate environment mechanics \cite{ha2018world}, MoL-RL relies solely on the LLM’s internal representations to interpret EF, potentially struggling with nonstationary environments. Future work could integrate world models to provide structured environmental predictions, complementing MoL-RL’s abstract reasoning with grounded dynamics.

\bibliography{sample}

\appendix
\onecolumn
\section{MoL training Example}
\noindent 
\refstepcounter{algorithm}
\begin{tcolorbox}[
    colback=gray!10,
    colframe=black,
    coltitle=white,
    colbacktitle=black,
    fonttitle=\bfseries,
    title=Example,
    boxrule=0.5mm,
    width=\textwidth,
    before=\medskip,
    after=\medskip,
]
[\\
\hspace*{4ex}\{\\
\hspace*{8ex}"role": "system",\\
\hspace*{8ex}"content": "You are an expert assistant who can solve any task using code blobs...",\\
\hspace*{4ex}\},\\
\hspace*{4ex}\{\\
\hspace*{8ex}"role": "user",\\
\hspace*{8ex}"content": "New task:
If $\log_6 (x-y) + \log_6 (x+y) = 2$ and $\log_y 5x = 2$, then find $x$. \\\hspace*{4ex}Write code, not latex.",\\
\hspace*{4ex}\},\\
\hspace*{4ex}\{\\
\hspace*{8ex}"role": "assistant",\\
\hspace*{8ex}"content": "<think>\\

\hspace*{4ex}</think>\\
\hspace*{4ex}Let's solve this step by step. We have two equations with logarithms:\\
\hspace*{4ex}1. $log_6(x-y) + log_6(x+y) = 2$\\
\hspace*{4ex}2. $log_y(5x) = 2$\\

\hspace*{4ex}...\\

\hspace*{4ex}Let me implement this step by step. "code": "\\
\hspace*{4ex}import math\\
\hspace*{4ex}import sympy as sp\\

\hspace*{4ex}\# Define variables\\
\hspace*{4ex}x, y = sp.symbols('x y')\\

\hspace*{4ex}\# Create equations\\
\hspace*{4ex}eq1 = sp.Eq(x**2 - y**2, 36)  \# from $log_6(x-y) + log_6(x+y) = 2$\\
\hspace*{4ex}eq2 = sp.Eq(5*x, y**2)        \# from $log_y(5x) = 2$\\

\hspace*{4ex}\# Solve the system of equations\\
\hspace*{4ex}solution = sp.solve((eq1, eq2), (x, y))\\
\hspace*{4ex}print(solution)"\\
\hspace*{4ex}\}\\
\hspace*{4ex}\{\\
\hspace*{8ex}"role": "user",\\
\hspace*{8ex}"content": "Observation:
Execution logs:
[(-4, -2*sqrt(5)*I), (-4, 2*sqrt(5)*I), (9, -\\\hspace*{4ex}3*sqrt(5)), (9, 3*sqrt(5))]
Last output from code snippet:
None",\\
\hspace*{4ex}\{\\
\hspace*{8ex}"role": "assistant",\\
\hspace*{8ex}"content": "..."\\
\hspace*{4ex}\},\\
\hspace*{4ex}...\\
]
\label{MoL_example}
\end{tcolorbox}

\section{MoL Continual Training Hyperparameters}
\label{MoL_Hyperparameters}
\begin{table}[h]
\centering
\label{tab:grpo training_setup}
\begin{tabular}{lc}
\toprule
\textbf{Hyperparameter} & \textbf{Value} \\
\midrule
global\_batch\_size & 72 \\
per\_device\_batch\_size & 3 \\
num\_devices & 3 (H20s) \\
gradient\_accumulation\_steps & 8 \\
learning\_rate& 1e-4 \\
LoRA\_rank & 64 \\
weight\_decay & 0.1 \\

\bottomrule
\end{tabular}
\caption{MoL continual training Training Setup}
\end{table}

\section{GRPO-based Post-Training Hyperparameters}
\label{GRPO_Hyperparameters}

\begin{table}[h]
\centering
\label{tab:grpo training_setup}
\begin{tabular}{lc}
\toprule
\textbf{Hyperparameter} & \textbf{Value} \\
\midrule
global\_batch\_size & 128 \\
per\_device\_batch\_size & 1 \\
num\_devices & 8 (H20s) \\
gradient\_accumulation\_steps & 16 \\
\bottomrule
\end{tabular}
\caption{Training Setup}
\end{table}

\begin{table}[h]
\centering

\label{tab:grpo rollout_config}
\begin{tabular}{lc}
\toprule
\textbf{Parameter} & \textbf{Value} \\
\midrule
rollout\_backend & vLLM \\
tensor\_parallel\_size & 1 \\
data\_parallel\_size & 1 \\
num\_generation & 8 \\
max\_completion\_length & 8192 \\
temperature & 0.9 \\
\bottomrule
\end{tabular}
\caption{Rollout Configuration}
\end{table}

\end{document}